\documentclass[letterpaper, 12pt]{article}

\usepackage{fancyhdr}

\fancypagestyle{firstpage}{
  \fancyhf{}
  \fancyhead[C]{This paper has been accepted for publication at the 2026 International Symposium on Medical Robotics (ISMR), Knoxville, Tennessee, USA.}
  
}

\usepackage{blindtext}
\usepackage{times} 
\usepackage{graphicx} 
\usepackage{bm} 
\usepackage{amsmath} 
\usepackage{amssymb} 
\usepackage{color} 
\usepackage[ruled,linesnumbered]{algorithm2e} 
\usepackage[short]{optidef} 
\usepackage{multirow}
\usepackage[mathscr]{eucal}
\usepackage{threeparttable}
\usepackage{booktabs}
\usepackage[
    backend=biber,
    style=ieee,
]{biblatex}
\newcommand{\homogtran}[2]{^{#2}T_{#1}}

\usepackage{xparse,soul} 
\soulregister\cite7 
\soulregister\citep7 
\soulregister\citet7 
\soulregister\ref7 
\soulregister\pageref7 
\usepackage{makecell}

\usepackage{caption}
\title{Comparative Analysis of Autonomous Robotic and Manual Techniques for Ultrasonic Sacral Osteotomy: A Preliminary Study}

\author{Daniyal Maroufi$^{1}$, Yash Kulkarni$^{1}$, Justin E. Bird$^{2}$, \\ Jeffrey H. Siewerdsen$^{3}$, and Farshid Alambeigi$^{1}$ {Member, IEEE}.
\thanks{This research is supported by the Collaborative Accelerator for Transformative Research Endeavors grant, jointly awarded by The University of Texas at Austin and The University of Texas MD Anderson Cancer Center. \\
$^{1}$D.~Maroufi, Y.Kulkarni, and F.~Alambeigi are with the Walker Department of Mechanical Engineering and the Texas Robotics  at the University of Texas at Austin, Austin, TX, 78712, USA. Email: maroufi@utexas.edu, kulkarni.yash08@utexas.edu,   farshid.alambeigi@austin.utexas.edu \\
$^{2}$ J. E. Bird is with the Department of Orthopedic Oncology, Division of Surgery, The University of Texas M.D. Anderson Cancer Center, Houston, TX, USA. Email: JEBird@mdanderson.org \\
$^{3}$ J. H. Siewerdsen is with the Department of Imaging Physics, Division of Diagnostic Imaging, The University of Texas MD Anderson Cancer Center, Houston, Texas, USA. Email: JHSiewerdsen@mdanderson.org}
}

\usepackage{color} 

\begin{document}
\date{}
\maketitle
\thispagestyle{firstpage}

\pagestyle{empty}
		
\begin{abstract}
In this paper, we introduce an autonomous Ultrasonic Sacral Osteotomy (USO) robotic system that integrates an ultrasonic osteotome with a seven-degree-of-freedom (DoF) robotic manipulator guided by an optical tracking system. To assess multi-directional control along both the surface trajectory and cutting depth of this system, we conducted quantitative comparisons between manual USO (MUSO) and robotic USO (RUSO) in Sawbones phantoms under identical osteotomy conditions. The RUSO system achieved sub-millimeter trajectory accuracy (0.11 mm RMSE), an order of magnitude improvement over MUSO (1.10 mm RMSE). Moreover, MUSO trials showed substantial over-penetration (16.0 mm achieved vs. 8.0 mm target), whereas the RUSO system maintained precise depth control (8.1 mm). These results demonstrate that robotic procedures can effectively overcome the critical limitations of manual osteotomy, establishing a foundation for safer and more precise sacral resections.
\end{abstract}

\section{Introduction}

Sacral tumors such as chordoma and chondrosarcoma are typically resistant to chemotherapy and radiation therapy. Therefore, the only curative treatment for these tumors  is often an en bloc sacrectomy, in which the tumor is excised along with a margin of healthy bone through a precisely planned osteotomy \cite{puri2009decision}. However, this manual procedure poses significant technical challenges because the osteotomy path lies in close proximity to the S1–S4 sacral nerve roots, which are essential for bowel, bladder, and sexual function \cite{hulen2006oncologic}. Even slight deviations from the planned trajectory can result in incomplete tumor resection or irreversible nerve injury. The procedure also carries high risks of massive blood loss and infection, further underscoring the critical need for technologies that can perform osteotomies with sub-millimeter accuracy and minimal  tissue damage \cite{tang2009risk}.

To address the above-mentioned challenges, computer-assisted navigation (CAN) has been introduced to improve the accuracy of manual osteotomy procedures. For example, Bosma et al. \cite{bosma2019can} reported that CAN-guided resections for primary pelvic and sacral sarcomas achieved an 81\% rate of negative margins compared with 50\% in conventional manual surgeries. Although these results highlight the benefits of CAN, the osteotomy is still performed manually by the surgeon, whose ability to precisely follow the planned trajectory and maintain the desired cutting depth is inherently limited. This manual execution increases the risk of cutting beyond the intended boundary or damaging critical underlying structures. Consequently, despite the significance of CAN, accuracy remains highly operator-dependent, and substantial deviations from the surgical plan continue to occur.
\begin{figure}
    \centering
    \includegraphics[width=0.9\linewidth]{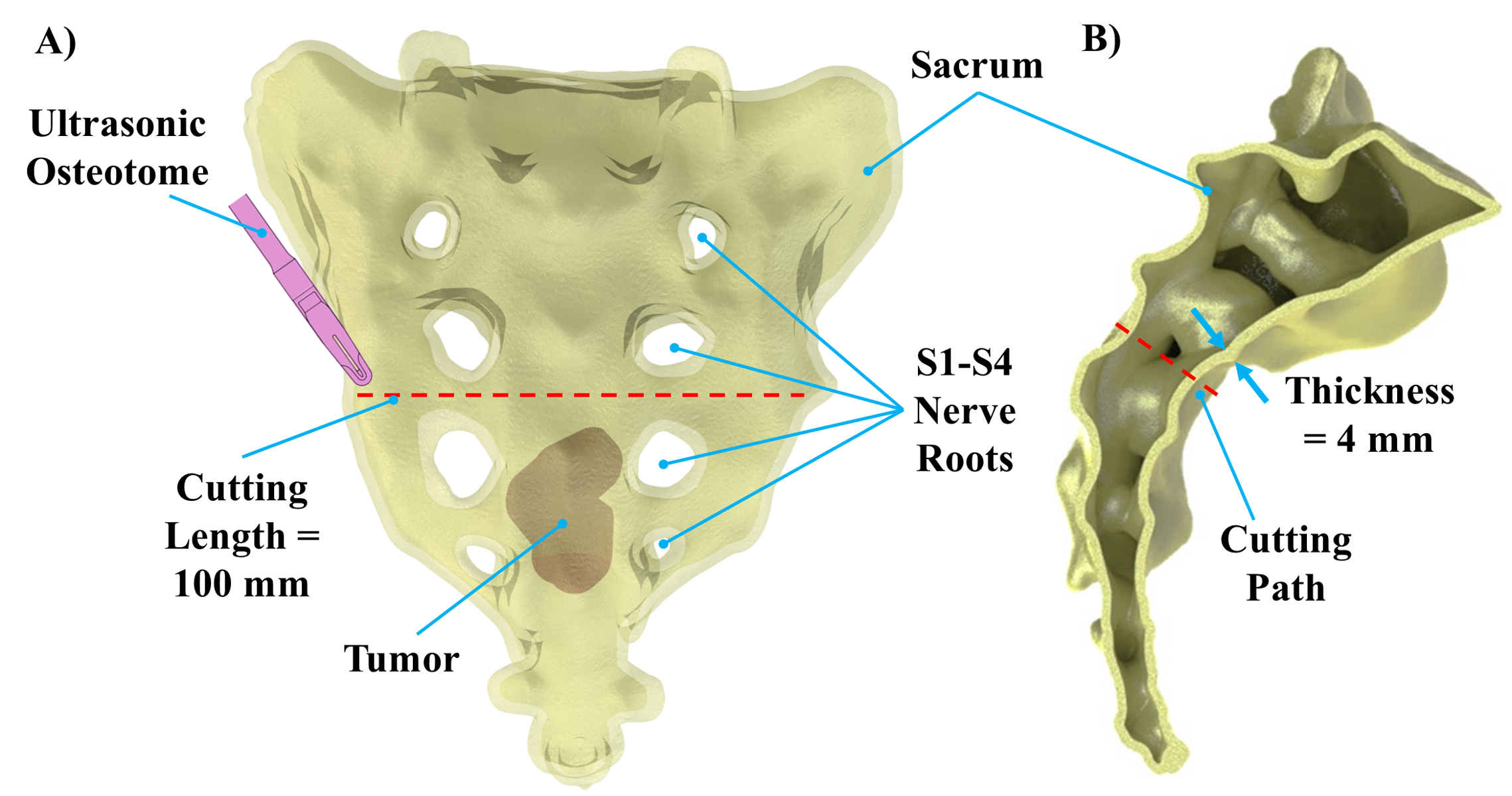}
    \caption{A) Anterior view of the sacrum illustrating the clinically-derived osteotomy task. The dashed line represents a 100 mm transverse cut path. B) Sagittal view of the sacrum.}
    \label{fig:concept}
\end{figure}
Across multiple studies, CAN-based sacral and pelvic osteotomies exhibit mean execution errors of 2--4~mm relative to the planned path~\cite{ritacco2013accuracy, sternheim2015navigated}, with a global mean deviation of 2.52~mm reported in clinical cases~\cite{ritacco2013accuracy}. More importantly, large and unpredictable outliers persist as reported by Ritacco \textit{et al.} ~\cite{ritacco2013accuracy}. In this study, authors have documented maximum deviations up to 18.5~mm and, critically, tumor-directed errors approaching 20~mm. Such variability, even with navigation assistance, primarily arises from the manual handling of osteotomes in a complex anatomical region like the sacrum, where bone geometry is highly variable and thin. Accurate osteotomy in this region demands precise multi-directional control, both along the surface trajectory and in cutting depth, yet maintaining such fine spatial coordination simultaneously can be very demanding even for expert clinicians. 
\begin{figure*}
    \centering
    \includegraphics[width=1.05\linewidth]{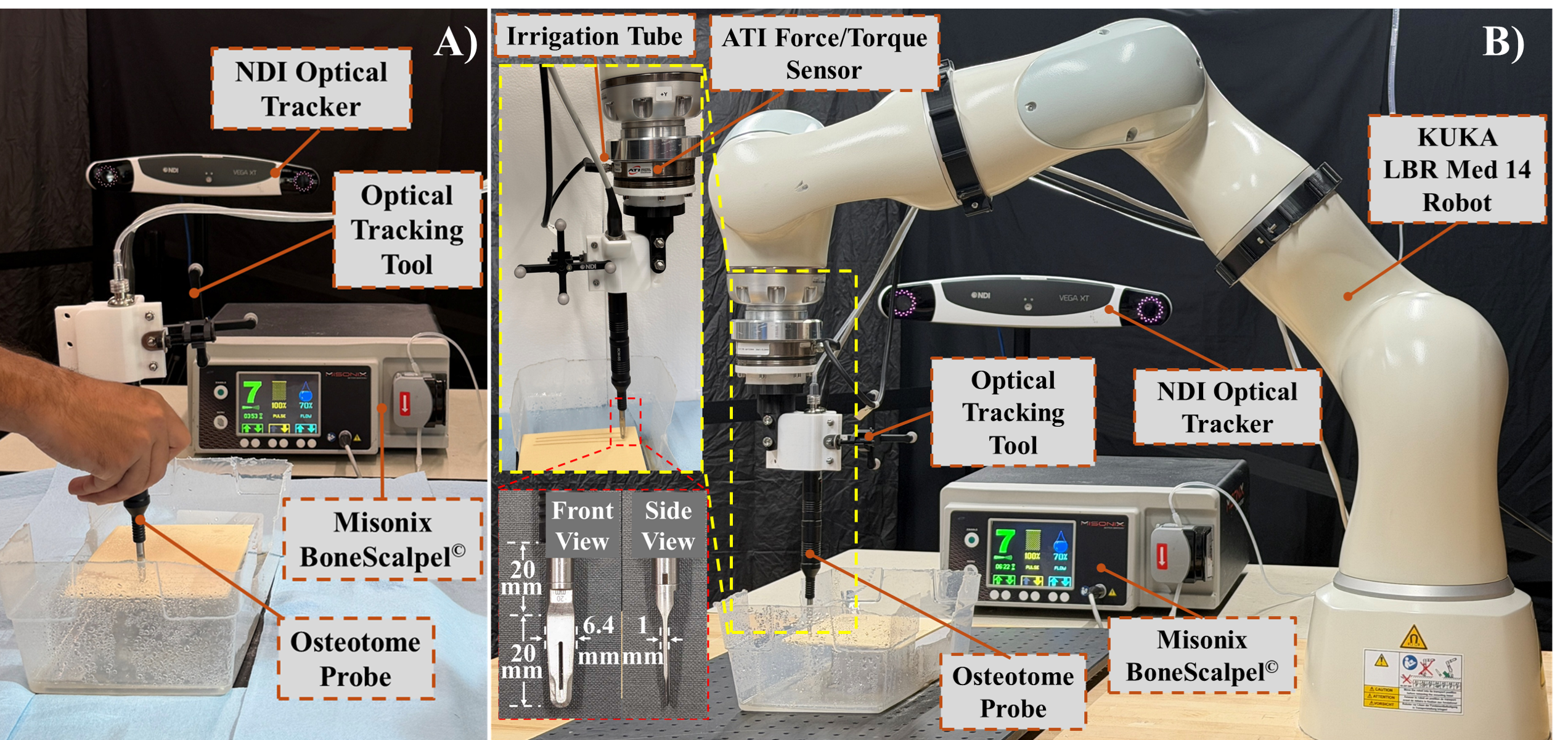}
    \caption{Experimental setup for the comparative experiments. A) The MUSO setup consisting of Misonix BoneScalpel and NDI Optical Tracker. B) The RUSO setup, featuring the KUKA LBR Med 14 Robot actuating the same osteotome.}
    \label{fig:setup}
\end{figure*}

To address the challenges asscoiated with CAN-based osteotomy, several researchers have used various robotic osteotomy procedures on different bony anatomies. For example, Bhagvath \textit{et al.} \cite{bhagvath2023design} developed an image-guided robotic spine osteotomy system. Using a milling machine as the osteotome, they performed a laminectomy procedure on a 3D printed vertebrae. In 2015, Baek \textit{et al.} developed a  laser osteotomy robotic system for cranio-maxillo-facial surgery \cite{baek2015clinical}. Authors used  an Er:YAG laser source as the osteotome to create defects of different shapes in the mandibles of 6 minipigs. More recently, a cold ablation robot-guided laser osteotome has also been introduced to improve the accuracy of interdental osteotomies in orthognathic surgery \cite{larsen2025robotic}.



To address the aforementioned challenges in performing en bloc sacrectomy, and as our main contribution, we introduce an autonomous Ultrasonic Sacral Osteotomy (USO) robotic system. The proposed system, shown in Fig. \ref{fig:setup}, integrates an ultrasonic osteotome with a seven-degree-of-freedom (DoF) robotic manipulator navigated using an optical tracking system. To evaluate multi-directional control along the surface trajectory and in cutting depth, we conduct a quantitative comparison between manual USO (MUSO) and autonomous robotic USO (RUSO) using Sawbones phantoms under identical osteotomy conditions. During these experiments, we record the three-dimensional cutting trajectories executed by an experienced human operator and by the autonomous robotic system. Key performance metrics, including trajectory accuracy, procedure time, and, most critically, achieved cut depth, are quantified and compared to assess a robotic sacrectomy procedure.


\section{Robotic Ultrasonic Sacral Osteotomy (RUSO) System}

As shown in Fig.~\ref{fig:setup}, our autonomous RUSO framework \cite{maroufi2026towards} is composed of several key components: (i) a seven-degree-of-freedom (DoF) robotic arm (KUKA LBR Med 14, KUKA, Germany) to position the ultrasonic osteotome to the target pose; (ii) an optical tracking system (NDI Polaris Vega, Northern Digital Inc.) employed for the calibration procedures, and to record the 3D trajectory of the tool for subsequent comparative analysis; and (iii) an ultrasonic osteotome (BoneScalpel, MISONIX LLC). 
The proposed procedure to perform a robotic sacrectomy using our RUSO system begins with a series of calibrations to establish the system's coordinate frames as described in Section \ref{sec:calibration}. 

\section{System Calibration Procedure}
\label{sec:calibration}

To facilitate a quantitative comparison between the manual and robotic osteotomies, all components of the experimental setup must be localized within a common coordinate system. In this work, we define the robot's base frame as $\{S\}$ and shown in Fig.~\ref{fig:transformations}, as the global reference frame. The spatial relationship between any two coordinate frames, for instance, frame $\{A\}$ and frame $\{B\}$, is a rigid body motion described as an element of the Special Euclidean group, $SE(3)$. This pose is represented by the homogeneous transformation matrix $\homogtran{A}{B}$, which defines the pose of frame $\{A\}$ relative to frame $\{B\}$. This section details the calibration and registration procedures required to compute the transformations that relate the robot base $\{S\}$ to the optical tracking system, the surgical tool, and the bone phantom, as illustrated in Fig.~\ref{fig:transformations}.

\subsection{Hand-Eye Calibration}
\label{sec:handeye}

\begin{figure}
    \centering
    \includegraphics[width=\linewidth]{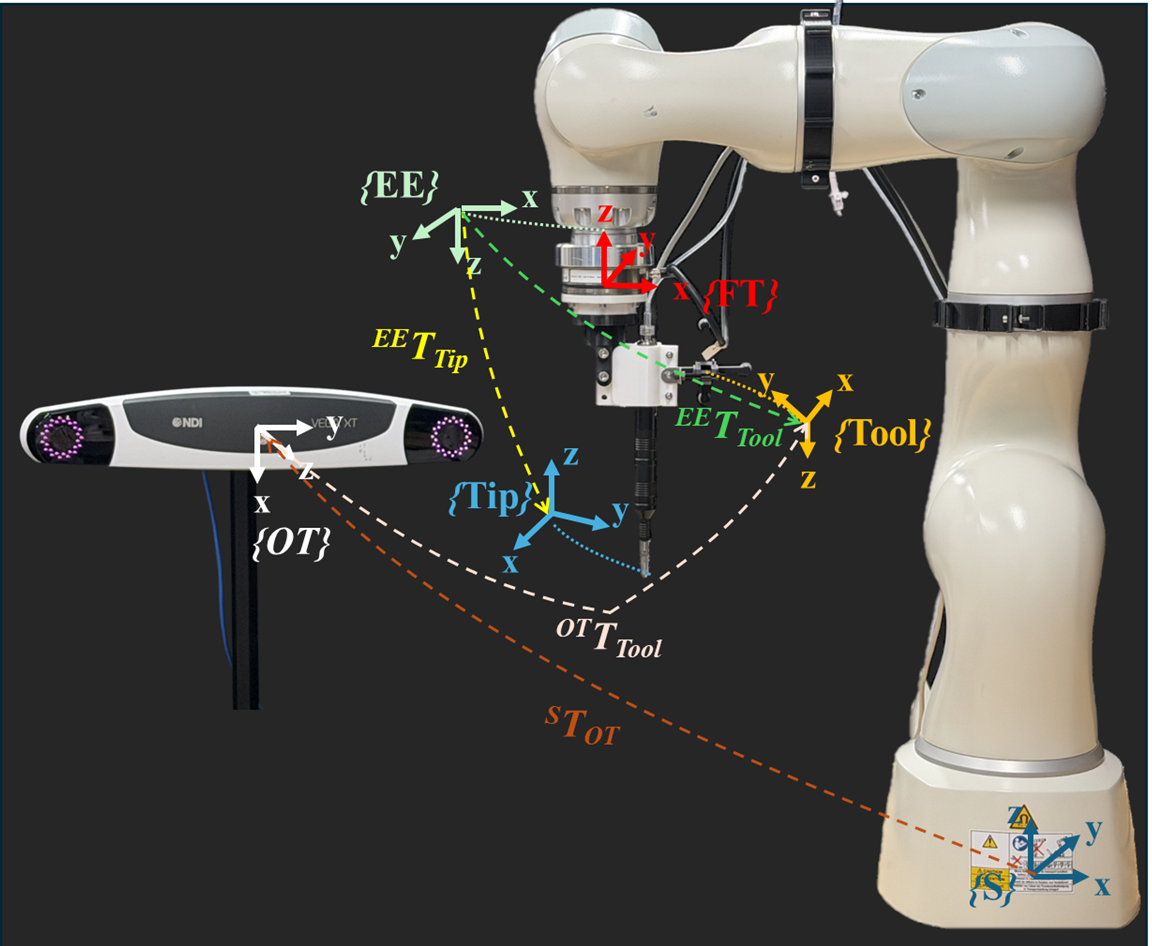}
    \caption{Coordinate frames and transformations for calibration and control. Frames assigned to key components of the system: robot base $\{S\}$, end-effector $\{EE\}$, osteotome tip $\{\text{Tip}\}$, optical tracker $\{OT\}$,  and the rigid body tool $\{\text{Tool}\}$. Key transformations, such as the hand-eye calibration ($\homogtran{OT}{S}$), are also shown.}
    \label{fig:transformations}
\end{figure}

The initial step in the registration process is to perform a hand-eye calibration to determine two critical transformations, as illustrated in Fig.~\ref{fig:transformations}. The first is the transformation from the optical tracker frame $\{OT\}$ to the robot base frame $\{S\}$, denoted as $\homogtran{OT}{S}$. The second is the transformation from the robot's end-effector frame $\{EE\}$ to the frame of the rigidly attached optical tracking tool $\{\text{Tool}\}$, denoted as $\homogtran{\text{Tool}}{EE}$.
To solve for these transformations, we employ a method based on the $AX=XB$ matrix equation \cite{shah2013solving}. The optical tracking tool $\{\text{Tool}\}$ was securely mounted to the robot's end-effector. The robot was then programmed to move to $N_{\text{handEye}}$ distinct poses distributed throughout its workspace. At each pose $i$, we simultaneously recorded the transformation from the end-effector to the robot base, $\homogtran{EE_i}{S}$, provided by the robot's internal kinematics, and the transformation from the optical tracking tool to the optical tracker's frame, $\homogtran{\text{Tool}_i}{OT}$, measured by the optical tracker system.

These measurements are related by the following kinematic loop closure equation:
\begin{equation}
    \homogtran{\text{Tool}_i}{OT} = \homogtran{S}{OT} \,\,.\,
    \homogtran{EE_i}{S} \,\,.\,
    \homogtran{\text{Tool}}{EE}
    \label{eq:kinematic_loop}
\end{equation}
By considering the relative motion between any two poses $i$ and $j$, this equation can be rearranged into the form $A_{ij}Y = YB_{ij}$, where $Y = {\homogtran{S}{OT}}$ is the unknown transformation between the robot base and the optical tracker. The matrices $A_{ij} = {\homogtran{EE_j}{S}}\,.\,(\homogtran{EE_i}{S})^{-1}$ and $B_{ij} = {\homogtran{\text{Tool}_j}{OT}}\,.\,(\homogtran{\text{Tool}_i}{OT})^{-1}$ represent the relative motion of the end-effector and the optical tracking tool, respectively. Using the collected set of $N$ poses, this overdetermined system is solved for $Y$.
Once $\homogtran{S}{OT}$ is determined, Equation~\ref{eq:kinematic_loop} can be reformulated as a linear system of equations of the form $A_i'X = B_i'$ to solve for the remaining unknown, $X = {\homogtran{\text{Tool}}{EE}}$. For each pose $i$, the matrices are defined as $A_i' = {\homogtran{EE_i}{S}}$ and $B_i' = (\homogtran{S}{OT})^{-1}\,.\,\homogtran{\text{Tool}_i}{OT}$. This procedure provides the complete set of transformations required to express the position and orientation of the robot and its tool within the global reference frame $\{S\}$.

\begin{figure}
    \centering
    \includegraphics[width=0.7\linewidth]{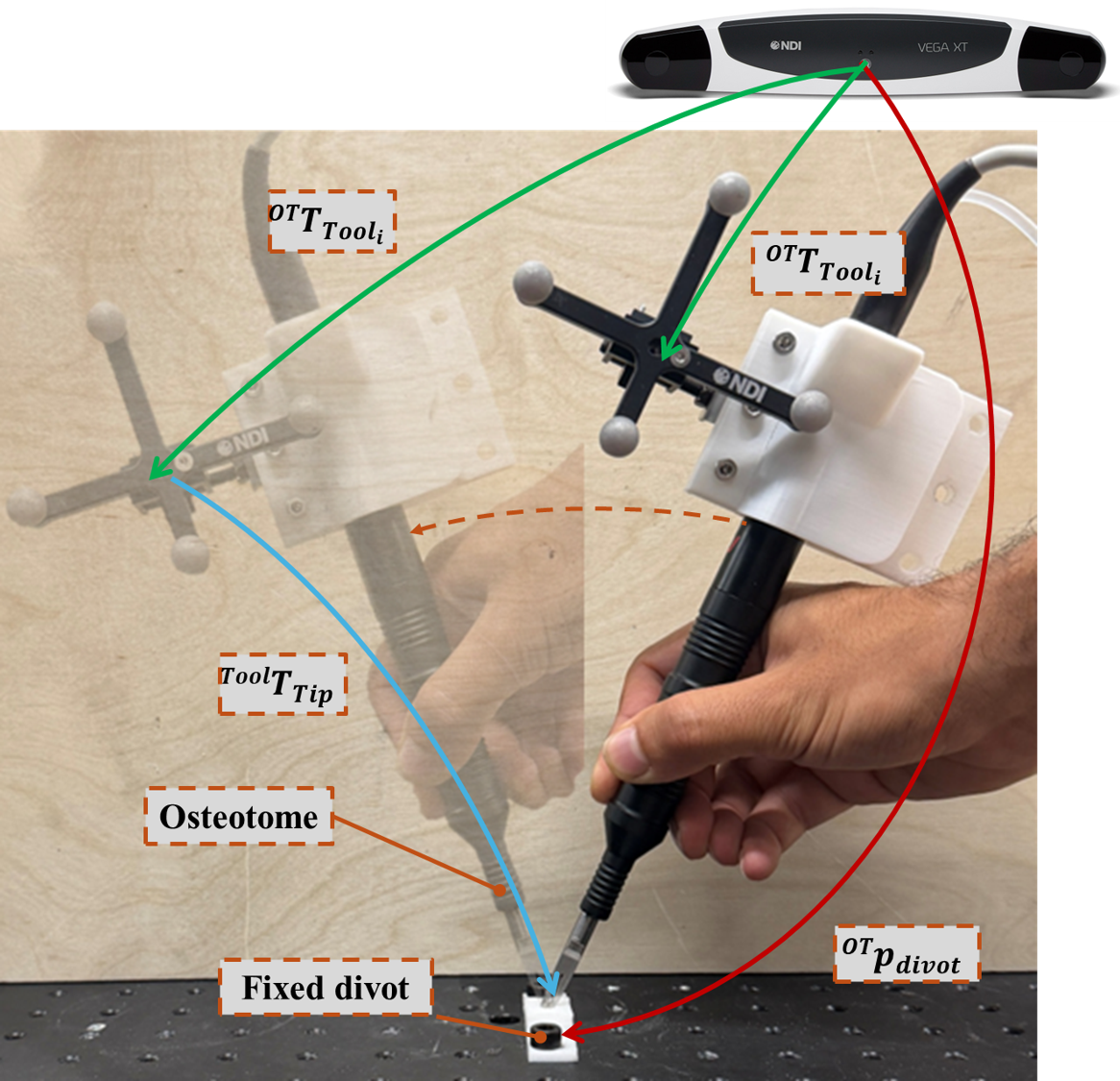}
    \caption{Pivot calibration process for localizing the osteotome tip. The optical tracker measures the transformation to the tool's rigid body ($^{OT}T_{\text{Tool}_i}$) across multiple poses as the tip is pivoted in a `Fixed divot`.}
    \label{fig:pivot}
\end{figure}

\subsection{Pivot Calibration}

To track the tip of the osteotome for MUSO procedures, a pivot calibration procedure was performed to determine the transformation from the optical tracking tool frame $\{\text{Tool}\}$ to the tip of the osteotome, denoted as the $\{\text{Tip}\}$ frame (Fig.~\ref{fig:transformations}). This transformation, $\homogtran{\text{Tip}}{\text{Tool}}$, is illustrated in Fig.~\ref{fig:pivot}. The calibration was conducted by placing the tip of the osteotome into a fixed divot (${}^{\text{OT}}\mathbf{p}_{\text{divot}}$), creating a stationary pivot point. While keeping the tip fixed, the body of the osteotome was pivoted through a wide range of orientations (as depicted in Fig.~\ref{fig:pivot}). During this motion, the optical tracker continuously recorded a set of $N_{\text{pivot}}$ poses of the tracking tool frame with respect to the tracker's base frame ($\homogtran{\text{Tool}_i}{OT}$, Fig.~\ref{fig:pivot}).

\begin{equation}
    {}^{\text{OT}}\mathbf{p}_{\text{divot}} = {\homogtran{\text{Tool}_i}{OT}} \,.\, {}^{\text{Tool}}\mathbf{p}_{\text{Tip}}
\end{equation}
where ${}^{\text{Tool}}\mathbf{p}_{\text{Tip}}$ is the translation part of $\homogtran{\text{Tip}}{\text{Tool}}$. By collecting a sufficient number of poses, a pivot calibration method as formulated in \cite{yaniv2015pivot} is conducted to obtain ${}^{\text{Tool}}\mathbf{p}_{\text{Tip}}$. 
This resulting vector defines the translational component of $\homogtran{\text{Tip}}{\text{Tool}}$, enabling precise tracking of the osteotome tip for subsequent experimental procedures.

\subsection{Tip Calibration}

To define the transformation of the osteotome tip with respect to the robot's end-effector ($\homogtran{\text{Tip}}{EE}$) for RUSO procedure, a tip calibration procedure is performed \cite{maroufi2025s3d}. In this method, an optical digitizer is used to point at the tip of the osteotome while holding different orientations. Using the Transformation obtained in \ref{sec:handeye}, and the transformation from the robot's end-effector to it's base frame ($\homogtran{EE}{S}$), $\homogtran{\text{Tip}}{EE}$ can be calculated:

\begin{equation}
    \homogtran{\text{Tip}}{EE} = (\homogtran{EE}{S})^{-1} \,.\,  \homogtran{OT}{S} \,.\, \homogtran{Digitizer}{OT}
\end{equation}

\section{Experimental Evaluations and Results}
\label{sec:experiments}


\subsection{Experimental Setup}

Figure \ref{fig:setup} shows the experimental setup used for our comparative studies including (i) the MUSO setup consisting of Misonix BoneScalpel with a length of 20 mm, 6.4 mm width, and 1 mm thickness, (ii) the RUSO system integrated with the ultrasonic osteotome,  (iii)  an optical tracking system (NDI Polaris Vega, Northern Digital Inc.); and (iv) a solid rigid polyurethane foam block (PCF 15 Sawbones, Pacific Research Laboratories, USA) used for replicating the sacrum hard tissue and conducting experiments.  Using this setup, the following experimental protocols were designed to directly compare the outcomes of the MUSO and RUSO techniques executing the sacrectomy.

 \subsection{Experimental Procedure}
 \label{sec:procedure}
To establish a standardized and clinically relevant basis for comparing the RUSO and MUSO procedures, a specific osteotomy task representative of a high sacral resection was defined. Figure~\ref{fig:concept} shows the target location, straight trajectory with 100 mm cut length for the planned osteotomy experiments. This trajectory was selected  based on anatomical studies of adult sacra, which report a mean sacral width ranging from approximately 84 mm to 126 mm \cite{katsuura2018anatomic}. To investigate performance across different difficulties, two target cutting depths were established for our experiments: 4 mm and 8 mm. Based on literature values indicating the anterior sacral cortical thickness \cite{richards2010bone}, the 4 mm depth was chosen to represent a complete and consistent transection of the cortical shell. The 8 mm depth was added to represent a more challenging task that involves cutting through both the cortical and underlying cancellous bone structures. These two depths allow for a comprehensive comparison of the techniques under different bone thicknesses. Before starting the cutting experiments, and based on the company’s recommended settings, the BoneScalpel was configured with amplitude level 7, pulse at 100\%, and water flow at 70\%. 

The designed experiments are listed in Table \ref{table:results} where M is used for MUSO experiments and R for RUSO. Since multiple trials are performed, to refer to a certain trial a super-scripted number is used throughout the paper (i.e. M1$^4$ is the forth trial from the M1 experiment set defined in Table \ref{table:results}). The experimental procedure for MUSO and RUSO experiments are as follows:

\subsubsection{MUSO experiments}

As shown in Fig. \ref{fig:setup}A, the manual osteotomy trials were performed to establish a performance baseline representing the current clinical standard of care. A trained and experienced operator conducted the procedures using an ultrasonic osteotome. To enable motion tracking, an NDI optical rigid body was attached to the osteotome using a custom-designed, 3D-printed (using a Raise3D 3D printer, Raise3D Technologies Inc. and PLA material) holder. This holder was designed to provide a stable mount without obstructing the operator's grip or field of view and was the same one used in the subsequent robotic trials to ensure experimental consistency.
To simulate a realistic surgical posture, the operator was standing, with the bone phantom securely fixed on a workbench. Following the predefined surgical plan visualized on the phantom's surface, the operator's objective was to execute the osteotomy task by cutting through the Sawbones blocks. The operator was permitted to perform multiple cutting passes to follow the desired straight trajectories while  trying  to not surpass the target depth (i.e., 4 mm and 8 mm). To ensure statistical reliability, the procedure was repeated four times, and the resulting data were averaged for the final analysis. 
Figures~\ref{fig:trajectories}B and~\ref{fig:trajectories}E illustrate representative cut depths from one of the trials in Sawbones phantoms for target depths of 4~mm and 8~mm, respectively.  The corresponding surface trajectories are shown in Figures~\ref{fig:trajectories}C and~\ref{fig:trajectories}F. The target depths are indicated by yellow boxes, and the desired thicknesses are shown by green dashed lines in these figures. Representative views of the resulting cuts are provided in Fig.~\ref{fig:cuts}.

\subsubsection{RUSO Experiments}

For the robotic experiments, we implemented an autonomous framework designed to constrain the motion of the surgical tool to the predefined osteotomy straight trajectories. This approach leverages the robot's precision to maintain trajectory and depth while the operator initiates and supervises the procedure.The procedure began with registering the physical bone phantom to the robot's workspace. An operator used an optically tracked digitizer to touch the starting point of the planned trajectory marked on the phantom's surface. The coordinates of this point were captured and transformed into the robot's base frame $\{S\}$, serving as the entry point for the automated cutting sequence.
Once registered, the robot executed the osteotomy by repeating the following automated sequence in increments until the full 3~mm depth was achieved: (1) \textit{Insertion}: The robot advanced the osteotome into the phantom to a predefined depth (i.e., 4 mm and 8 mm) at a controlled insertion speed; (2) \textit{Cutting}: At the target depth, the robot translated the osteotome along the planned 100~mm straight trajectory at a constant cutting speed; and (3) \textit{Retraction}: After completing the pass, the robot retracted the tool from the bone.
Throughout this osteotomy procedure, the optical tracking system continuously recorded the executed tool trajectory for subsequent analysis. For each experiment set, three trials were performed on bone phantoms to ensure statistical reliability. Figures~\ref{fig:trajectories}A and~\ref{fig:trajectories}D illustrate the achieved cut depths in Sawbones phantoms for target depths of 4~mm and 8~mm, respectively. Corresponding surface views of these cuts are shown in Figure~\ref{fig:cuts}.

\subsection{Evaluation Metrics}

To provide a comprehensive and quantitative comparison between the MUSO and RUSO techniques, the data recorded during each trial were analyzed using four primary metrics: trajectory accuracy, executed length, procedure time, and achieved depth. To formalize this comparison, we denote each trial as $\mathcal{X}i^j$, where $\mathcal{X} \in \{\text{M}, \text{R}\}$ indicates the technique (MUSO or RUSO, respectively), and $i$ denotes the set of experimental parameters as they follow the task defined in Section~\ref{sec:procedure}. The superscript $j$ represents the trial number for that set.

\subsubsection{Trajectory Accuracy}
Trajectory accuracy was evaluated by computing the deviation of the executed path from the ideal, planned trajectory. For each trial $\mathcal{X}i^j$, the executed path, as recorded by the optical tracker, consists of a set of $M$ 3D points, $\mathbf{P} = \{ \mathbf{p}_1, \dots, \mathbf{p}_M \}$, where each point $\mathbf{p}_k = [x_k, y_k, z_k]^T$ is expressed in the robot's base frame $\{S\}$. The planned trajectory was defined as an ideal straight line, $\mathcal{L}_{\text{plan}}$, in this same frame.
For each measured point $\mathbf{p}_k$, the perpendicular error, $e_k$, was calculated as its minimum Euclidean distance to the ideal line $\mathcal{L}_{\text{plan}}$. We use the Root Mean Squared Error (RMSE) as the primary metric for overall accuracy. The RMSE for the trial, $e_{\text{RMSE}}(\mathcal{X}i^j)$, is calculated as the square root of the mean of these squared errors, $e_{\text{RMSE}}(\mathcal{X}i^j) = \sqrt{\frac{1}{M} \sum_{k=1}^{M} e_k^2}$.

\subsubsection{Executed Length}
The final executed length of the osteotomy, $L(\mathcal{X}i^j)$, was determined by a direct physical measurement of the cut on the bone phantom. After each trial, a digital caliper was used to measure the total length of the cut on the phantom's surface. This physical measurement was recorded and compared to the target length of 100 mm to assess the completeness of the cut.

\subsubsection{Procedure Time}
The procedure time, $t(\mathcal{X}i^j)$, was defined as the total active tool time required to complete the osteotomy. This duration was measured by logging the time the Misonix BoneScalpel's ultrasonic system was actively powered on and cutting, from the start of the first pass to the completion of the final pass.

\subsubsection{Achieved Depth}
As shown in Fig. \ref{fig:trajectories}, evaluating the achieved depth is complex, as the manual procedure involves multiple cutting passes over the same trajectory. A simple average of all $z$-coordinates would be misleading. To accurately determine the final cut profile, we implemented an analysis to find the maximum depth at each point along the trajectory.
For each trial $\mathcal{X}i^j$, the 3D trajectory data $\mathbf{P} = \{ \mathbf{p}_k \}$ was first projected onto the 2D cutting plane (X-Z plane) to create a profile set $\mathbf{P}' = \{ (x_k, z_k) \}_{k=1}^M$. The 100 mm trajectory along the $x$-axis was then discretized into $K$ uniform bins (e.g., $K=100$ for 1 mm bins), where each bin $B_j$ corresponds to a specific segment of the cut.
A function $\mathcal{D}$ was used to compute the final cut profile for each trial $\mathcal{X}i^j$. The $j$-th element of this profile, $d_j$, is defined as the minimum $z$-value (representing the deepest point reached) from all trajectory points whose $x$-coordinate falls within that bin:
\begin{equation}
    d_j = \min \{ z_k \mid (x_k, z_k) \in \mathbf{P}' \land x_k \in B_j \}
\end{equation}
where $\land$ is logical AND operator. This resulting profile, denoted as $\mathcal{D}(\mathcal{X}i^j)$, represents the final shape of the osteotomy floor as demostrated in Fig.\ref{fig:trajectories}B, E. The overall mean achieved depth for the trial, $\bar{d}(\mathcal{X}i^j)$, was then calculated as the mean of this profile's elements, $\bar{d}(\mathcal{X}i^j) = \frac{1}{K} \sum_{j=1}^{K} d_j$. This mean value was compared against the target 4mm and 8 mm depth.  Table~\ref{table:results} summarizes and compares these quantitative metrics obtained from both RUSO and MUSO experiments.

\section{Discussion}
\label{sec:results}

\begin{figure}
    \centering
    \includegraphics[width=\linewidth]{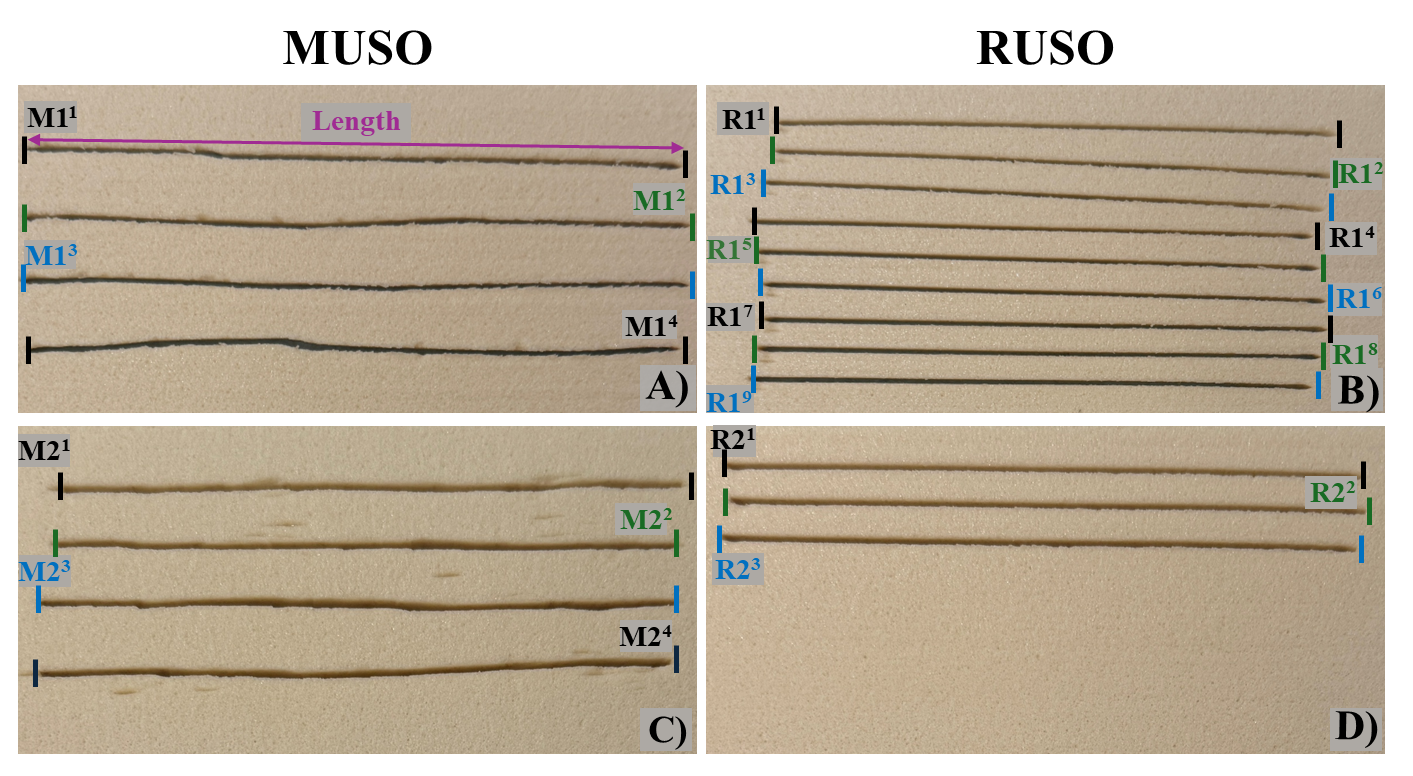}
    \caption{Qualitative comparison of the completed osteotomy cuts on the bone phantoms: A) MUSO trials and B) RUSO trials for the 4 mm target depth. C) MUSO trials and D) RUSO trials for the 8 mm target depth. The super-scripted numbers are the trial numbers. \\}
    \label{fig:cuts}
\end{figure}

\begin{figure}
    \centering
    \includegraphics[width=1\linewidth]{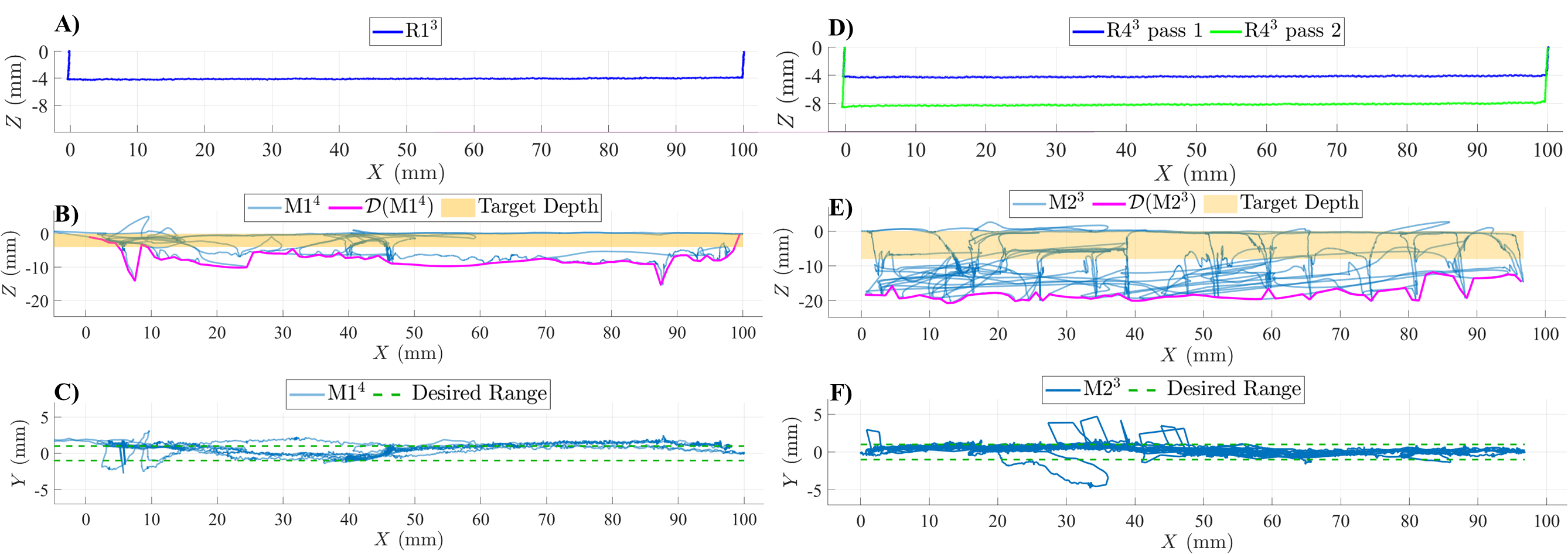}
    \caption{Representative trajectory plots for RUSO and MUSO experiments. A) RUSO side view for the 4 mm target depth (R1$^3$, third trial of R1 as defined in Table \ref{table:results} and shown in Fig. \ref{fig:cuts}). B) MUSO side view for the 4 mm target depth (M1$^4$), and the achieved depth $\mathcal{D}(\text{M1}^4)$. C) MUSO top view for the 4 mm target depth (M1$^4$). D) RUSO side view for the 8 mm target (R4$^3$), showing two controlled passes. E) MUSO side view for the 8 mm target depth (M2$^3$), and the achieved depth $\mathcal{D}(\text{M2}^3)$. F) MUSO top view for the 8 mm target depth (M2$^3$). \\ }
    \label{fig:trajectories}
\end{figure}

\begin{table*}[t]
\centering
\caption{Quantitative results comparing MUSO and RUSO experiments All values are reported as mean $\pm$ standard deviation over all trials for each experimental set. For RUSO, "Cutting Speed" was a controlled parameter, while for MUSO, it is the calculated average speed (length/time).}
\label{table:results}
\small
\setlength{\tabcolsep}{4pt}
\begin{tabular}{lcccccc}
\toprule
 & \textbf{Target} & \textbf{Cutting} & \textbf{RMSE} & \textbf{Length} & \textbf{Procedure} & \textbf{Depth} \\
 & \textbf{Depth} & \textbf{Speed} & \textbf{(mm)} & \textbf{(mm)} & \textbf{Time (s)} & \textbf{(mm)} \\
 & \textbf{(mm)} & \textbf{(mm/s)} & & & & \\
\midrule
M1 & 4 & 1.68 $\pm$ 0.15 & 1.10 $\pm$ 0.26 & 97.4 $\pm$ 1.2 & 58.25 $\pm$ 4.86 & 7.0 $\pm$ 0.8 \\
M2 & 8 & 0.91 $\pm$ 0.19 & 0.93 $\pm$ 0.21 & 97.9 $\pm$ 1.5 & 111.50 $\pm$ 27.16 & 15.9 $\pm$ 2.2 \\
\midrule
R1 & 4 & 1 & 0.13 $\pm$ 0.01 & \textbf{100.9 $\pm$ 0.2} & 127.33 $\pm$ 1.15 & 4.21 $\pm$ 0.12 \\
R2 & 4 & 2 & 0.11 $\pm$ 0.04 & 101.3 $\pm$ 0.2 & 64.33 $\pm$ 0.58 & 4.20 $\pm$ 0.16 \\
R3 & 4 & 3 & 0.10 $\pm$ 0.01 & 101.6 $\pm$ 0.1 & \textbf{44.33 $\pm$ 0.58} & 4.20 $\pm$ 0.04 \\
R4 & 8 & 3 & 0.09 $\pm$ 0.05 & 102.2 $\pm$ 0.1 & 90.33 $\pm$ 0.58 & 8.06 $\pm$ 0.01 \\
\bottomrule
\end{tabular}
\end{table*}


As summarized in Table \ref{table:results} and observed in Fig. \ref{fig:trajectories}, a clear performance gap is evident across all metrics, beginning with procedure time ($t$). The MUSO trials were significantly longer, with an average active cutting time of 111.3 s for the 4 mm depth task. This reflects the difficulty of the manual procedure, which requires multiple, cautious, and overlapping passes. In contrast, the RUSO procedure time was deterministic and controlled. As shown in Table~\ref{table:results}, we were able to execute the same task in as little as 45 s (at 3 mm/s) without any degradation in accuracy or depth control. This demonstrates a significant gain in efficiency.
This control is also reflected in the executed length ($L$) and trajectory accuracy ($e_{\text{RMSE}}$). The RUSO system consistently produced cuts of a reliable length, averaging 101.1 mm across all trials. The MUSO trials, however, were less consistent and failed to meet the 100 mm target, with one trial (M2$^2$) being as short as 96 mm. More critically, the trajectory accuracy of the MUSO trials was poor, with $e_{\text{RMSE}}$ values of 1.10 mm (M1) and 0.93 mm (M2). The RUSO system, by comparison, achieved a consistent sub-millimeter accuracy, with an average $e_{\text{RMSE}}$ of 0.11 mm, an order of magnitude better than the manual approach.

The most significant finding is the lack of depth control ($\bar{d}$) in the MUSO experiments. As seen in Table~\ref{table:results}, for the 4 mm target depth (M1), the operator achieved a mean depth of 7.0 mm (75\% over-penetration). This error was amplified with a deeper target, where the 8~mm task (M2) resulted in a 16 mm mean depth, doubling the intended cut depth. The side-view depth plots in Fig.~\ref{fig:trajectories}B, E clearly visualize this, showing deep, uncontrolled over-cuts. In contrast, the RUSO system (Fig.~\ref{fig:trajectories}A, D) demonstrated precise depth control, achieving mean depths of 4.2~mm and 8.1~mm, respectively, regardless of cutting speed. The over-cut depth  in the MUSO trials is clinically concerning, as they could penetrate through  critical nerve roots \cite{ritacco2013accuracy}.
As reported in Table~\ref{table:results}, for the 8 mm cuts, the manual trials required an average of 111 seconds (M2) corresponding to about 0.9 mm/s average speed, a duration necessitated by the multiple, overlapping, and uncontrolled passes visible in the trajectory plots. Nevertheless, the similar RUSO procedure, by executing stable passes, completed the task deterministically in 90 seconds (with 3 mm/s average speed) without compromising accuracy or depth. The  obtained trajectories in Fig. \ref{fig:trajectories} as well as post-operative photographs (Fig.~\ref{fig:cuts}) confirm these findings: the RUSO cuts are clean, straight, and uniform, whereas the MUSO cuts are visibly wider, jagged, and less consistent, reflecting the poor trajectory and depth control.
Overall, the MUSO experiments were an order of magnitude less accurate in trajectory (1.10 mm vs. 0.11 mm RMSE; Table~\ref{table:results}) and exhibited a critical lack of depth control, with 75–100\% over-penetration due to excessive manual force (Fig.~\ref{fig:trajectories}). In contrast, the RUSO system achieved precise, repeatable cuts with near-perfect depth control and consistent execution timing.

\section{Conclusion}
\label{sec:conclusion}

In this paper, we introduced an autonomous USO robotic system and evaluated its quantitative multi-directional control along the surface trajectory and in cutting depth using Sawbones phantoms. Comparative experiments were conducted between the RUSO and MUSO approaches. The RUSO system achieved sub-millimeter trajectory accuracy (0.11 mm RMSE) and executed cuts with precise and reliable depth control (less than 5\% error), completing each trajectory of length 100 mm in 45 seconds for a target depth of 4 mm. In contrast, the MUSO trials exhibited higher trajectory errors (1.1 mm) and, most critically, frequent and unsafe over-penetrations, reaching depths up to twice the intended target (75\% and 100\% for target depth of 4 mm and 8 mm, respectively). These findings provide empirical evidence that robotic automation can overcome key limitations of manual osteotomy execution by enhancing precision, ensuring depth control, and improving procedural efficiency. Future work will focus on extending this robotic capability to more complex, non-linear osteotomy trajectories designed to navigate around critical anatomical structures such as the sacroiliac joint \cite{Kulkarni2025SynergisticPSA,maroufi2026systematic,Kulkarni2026OFDRCTSDR,kulkarni2025sshape}.

\printbibliography

\end{document}